\documentclass{article}
\usepackage{url}            
\usepackage{booktabs}       
\usepackage{amsfonts}       
\usepackage{pifont}         
\usepackage{nicefrac}       
\usepackage{microtype}      
\usepackage{xcolor}         
\usepackage{graphicx}
\usepackage{multirow}
\usepackage{PRIMEarxiv}
\usepackage{pifont}
\usepackage[utf8]{inputenc} 
\usepackage[T1]{fontenc}    
\usepackage{hyperref}       
\usepackage{url}            
\usepackage{booktabs}       
\usepackage{amsfonts}       
\usepackage{nicefrac}       
\usepackage{microtype}      
\usepackage{lipsum}
\usepackage{fancyhdr}       
\usepackage{graphicx}       
\graphicspath{{media/}}
\usepackage{makecell}

\title{MCTBench: Multimodal Cognition towards Text-Rich Visual Scenes Benchmark
}

\author{
 \textnormal{
 \normalsize{Bin Shan\textsuperscript{\rm 1}\thanks{ Equal contribution.} , Xiang Fei\textsuperscript{\rm 1}$^{*}$, Wei Shi\textsuperscript{\rm 1}$^{*}$, An-Lan Wang\textsuperscript{\rm 1}$^{*}$,
Guozhi Tang\textsuperscript{\rm 1}, Lei Liao\textsuperscript{\rm 1}}, 
} 
\\
\normalsize{
Jingqun Tang \textsuperscript{\rm 1},Xiang Bai\textsuperscript{\rm 2},Can Huang\textsuperscript{\rm 1}
} 
\\
\textnormal{\small{\textsuperscript{\rm 1}ByteDance \textsuperscript{\rm 2}Huazhong University of Science and Technology}}
\\
\small{
\{shanbin, feixiang.77, shiwei.11, wanganlan, can.huang\}@bytedance.com 
}
}

\begin{document}

\maketitle

\begin{abstract}
The comprehension of text-rich visual scenes has become a focal point for evaluating Multi-modal Large Language Models (MLLMs) due to their widespread applications. Current benchmarks tailored to the scenario emphasize perceptual capabilities, while overlooking the assessment of cognitive abilities. To address this limitation, we introduce a \textbf{M}ultimodal benchmark towards \textbf{T}ext-rich visual scenes, to evaluate the \textbf{C}ognitive capabilities of MLLMs through visual reasoning and content-creation tasks (\textbf{MCTBench}). To mitigate potential evaluation bias from the varying distributions of datasets,  MCTBench incorporates several perception tasks (e.g., scene text recognition) to ensure a consistent comparison of both the cognitive and perceptual capabilities of MLLMs. To improve the efficiency and fairness of content-creation evaluation, we conduct an automatic evaluation pipeline. Evaluations of various MLLMs on MCTBench reveal that, despite their impressive perceptual capabilities, their cognition abilities require enhancement. We hope MCTBench will offer the community an efficient resource to explore and enhance cognitive capabilities towards text-rich visual scenes. 
\end{abstract}

\section{Introduction}
Multimodal Large Language Models (MLLMs) \cite{Achiam2023GPT4TR,team2023gemini,liu2023llava,li2024minigemini} have exhibited promising performance across various cross-modal tasks, and revealed potential for widespread real-world applications. In practical applications, many images contain crucial textual elements that are essential for addressing specific challenges, such as key information extraction from receipts.  Consequently, the ability to comprehend text-rich visual scenes can significantly enhance the practicality of MLLMs and drive innovative applications across multiple domains.

\begin{figure}
    \centering
    \includegraphics[width=\textwidth]{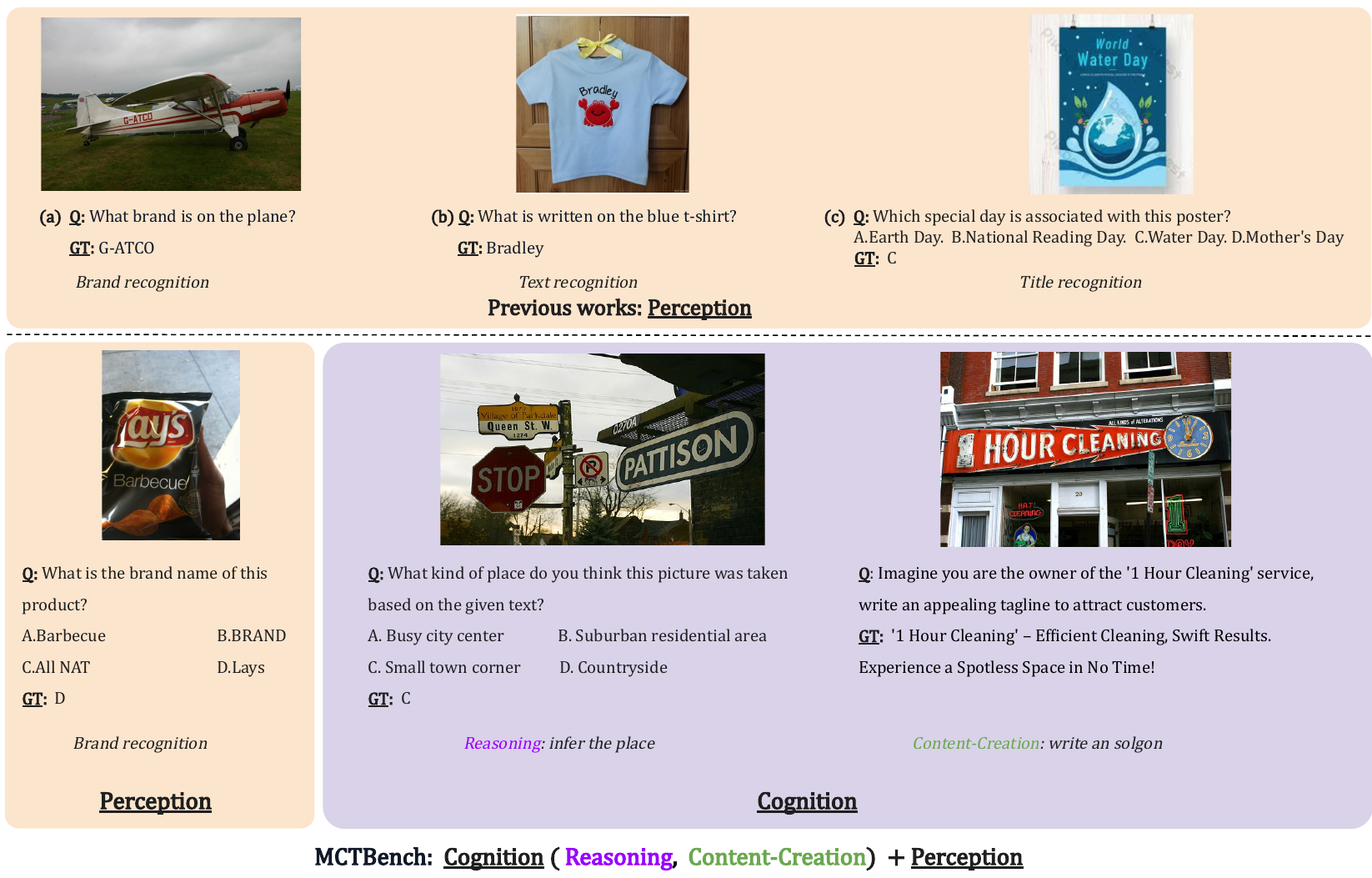}
    \caption{ The Comparison between previous Benchmarks \cite{Singh_2019_CVPR,liu2024hidden,li2024seedbench2plus}, and our proposed MCTBench. \textbf{Q} and \textbf{GT} stand for question and ground truth.}
    \label{fig:motivation}
\end{figure}

Recent benchmarks \cite{liu2024hidden, li2024seedbench2plus, yue2023mmmu} have increasingly focused on evaluating MLLMs towards text-rich visual scenes. Nonetheless, the benchmarks are centered around evaluating perceptual capabilities yet overlook the assessment of cognitive abilities, which are a significant strength of MLLMs (as illustrated in Figure \ref{fig:motivation}).

In this paper, we propose a \textbf{M}ultimodal benchmark to evaluate the \textbf{C}ognitive capabilities of MLLMs in \textbf{T}ext-rich visual scenes (\textbf{MCTBench}). To assess the cognitive abilities of MLLMs thoroughly, we design two types of tasks in the MCTBench: reasoning tasks for comprehension of the in
put scenes, and open-ended content-creation tasks for generating output responses. 
Besides, MCTBench integrates various perception tasks to study the differences with cognition tasks, while avoiding evaluation biases from varying dataset distributions. 
Fundamentally, MCTBench curates approximately 5.2k text-rich images from a wide range of public datasets, along with 8.5k rigorously annotated question-answer pairs categorized into three tasks: perception, reasoning and content-creation. The perception and reasoning tasks are formatted as multiple-choice questions for convenient evaluation, following common practices in \cite{fu2024mme,liu2024mmbench,li2023seedbench}. 
Due to the subjectivity and high cost of human evaluation in open-ended content creation, we establish an automated evaluation pipeline by leveraging sophisticated MLLMs (e.g., GPT-4V) as the evaluator, to compare the predictions of models against the provided references. Our experimental results demonstrate that MLLMs exhibit notably lower performance of cognition capabilities compared to perception in text-rich visual scenes, particularly for text-enhanced models. Furthermore, performances in cognition tasks (reasoning and content-creation) are improved with larger parameter scales.
Our main contributions are summarized as follows:
\begin{enumerate}
    \item We propose a brand-new and large-scale benchmark for evaluating the cognitive capability of MLLMs towards text-rich visual scenes.
    \item The evaluation on MCTBench highlights that MLLMs necessitate enhancements in their cognitive capabilities in text-rich visual scenes.
    \item We develop an automated evaluation pipeline for the content-creation task, offering researchers an efficient tool for further investigation of cognitive capabilities.
\end{enumerate}

\section{Related Work}
\subsection{Multimodal Large Language Models}
The significant advancements in Large Language Models (LLMs) \cite{Achiam2023GPT4TR, Touvron2023LLaMAOA, vicuna2023} have paved the way for recent research \cite{team2023gemini, bai2023qwen, liu2023llava, chen2023minigptv2, dai2023instructblip} into developing Multimodal Large Language Models (MLLMs) that integrate visual capabilities. Early works in this field \cite{alayrac2022flamingo, li2023blip2, liu2023llava, chen2023minigptv2} have introduced various vision-language projectors such as Q-former\cite{dai2023instructblip}, Multi-Layer Perceptron (MLP) \cite{liu2023llava}, and Perceiver\cite{alayrac2022flamingo}, which act as intermediaries between LLMs and visual encoders. Furthermore, these efforts have also established robust training paradigms for MLLMs. Building upon these foundational paradigms, recent initiatives \cite{chen2023sharegpt4v, lu2024deepseekvl, liu2023improvedllava, mckinzie2024mm1} have focused on scaling the quality of training data, to enhance general visual capabilities effectively.

A primary challenge in the recent development of MLLMs is attaining fine-grained comprehension, exemplified by tasks such as Visual Question Answering (VQA) on text-rich images.
To address this issue, increasing the resolution and integrating fine-grained visual features have been proven effective across various studies~\cite{feng2023unidoc, liu2024textmonkey, liu2024llavanext, Hu2023BLIVAAS, ye2023mplugowl2, ye2023mplugdocowl, li2023monkey, wei2023vary}. Additionally, works such as \cite{feng2023unidoc, Hu2023BLIVAAS, zhang2024llavar, li2023monkey, tang2024textsquare} have incorporated high-quality, text-rich visual tuning data to refine these models further.
\subsection{MLLM Benchmarks}
As multimodal large language models (MLLMs) continue to exhibit cross-task generality, single-task evaluations (e.g., \cite{balanced_vqa_v2, Singh_2019_CVPR, chen2015microsoft, lin2024rethinking}) are inadequate for a comprehensive performance assessment. Recent works \cite{liu2024mmbench, li2023seedbench, fu2024mme, yu2023mmvet} present general MLLM benchmarks comprising multiple tasks. Furthermore, to explore the performance of MLLMs on more complex tasks, MathVista \cite{lu2024mathvista} evaluates their mathematical abilities, and MMMU \cite{yue2023mmmu} integrates multiple-discipline questions to benchmark MLLMs in expert domains.

Conversely, text-rich visual scenes are attracting growing attention due to their potential applications. Early works \cite{mathew2021docvqa, mishra2019ocr, Singh_2019_CVPR} focused on single tasks, while OCRBench \cite{liu2024hidden} integrates multiple single-task datasets into five representative OCR(Optical Character Recognition)-based tasks. In contrast, our work evaluates MLLMs on complex tasks beyond OCR-based ones in text-rich visual scenes. A similar work is presented in \cite{wadhawan2024contextual}, which demonstrates the model's performance on reasoning tasks but only on a limited set of test datasets. Our study provides a broader evaluation of cognition in text-rich visual scenes, pushing the boundaries of what MLLMs can achieve in more diverse scenarios such as content-creation. Table \ref{related benchmark} demonstrates the detailed comparison between ours and previous benchmarks.

\begin{table}[]
\centering
\setlength\tabcolsep{0.5pt}
\begin{tabular}{lccccccc}         
\toprule
\multirow{1}{*}{Benchmark}  & {\makecell{Text-Rich\\Oriented}}  & \multirow{1}{*}{\#Image} & \multirow{1}{*}{\#QAs}  & \multirow{1}{*}{Percetion} & \multirow{1}{*}{Reasoning} & {\makecell{Content\\Creation}} & {\makecell{Answer\\Type}} \\
\midrule
MME \cite{fu2024mme}  & \ding{56} & 1137 &  2.2K  & \ding{52} & \ding{52} &  & Yes/No  \\
MMBench \cite{liu2024mmbench} & \ding{56}  &  -  & 3K  & \ding{52} & \ding{52} &   & MC  \\
\hline
OCRBench \cite{liu2024hidden}  & \ding{52} &   450   &  1K    & \ding{52} &  &  & Open   \\
SEED-bench-2-plus \cite{li2024seedbench2plus} & \ding{52} & - & 2.3K & \ding{52} &  &  & MC \\
Contextual \cite{wadhawan2024contextual}  & \ding{52} & 506 & 506  & \ding{52} & \ding{52} &  & Open   \\
MMMU \cite{yue2023mmmu}   & \ding{52}  & - & 11.5K & \ding{52} & \ding{52} &  &  MC/Open  \\
\textbf{MCTBench} & \ding{52} & 5.2K & 8.5K & \ding{52} & \ding{52} & \ding{52} & MC/Open     \\
\bottomrule
\end{tabular}
\caption{ The comparison between MCTBench and previous benchmarks. \textbf{Open} and \textbf{MC} respectively present open-ended and multiple choice format for answer type. \textbf{QAs} stands for question-answer pairs. Text-Rich Oriented indicates whether the benchmark focuses on text-rich visual scenes. }
\label{related benchmark}
\end{table}

\section{MCTBench}

In this section, we outline the process of constructing the MCTBench. Section \ref{overview} provides an overview of MCTBench and compares it with previous benchmarks. Section \ref{DataCollection} describes the procedure of collecting text-rich image sources from publicly accessible datasets. Finally, Section \ref{DataConstruction} explains the annotation process applied to the collected images.

\begin{figure}
    \centering
    \includegraphics[width=\textwidth]{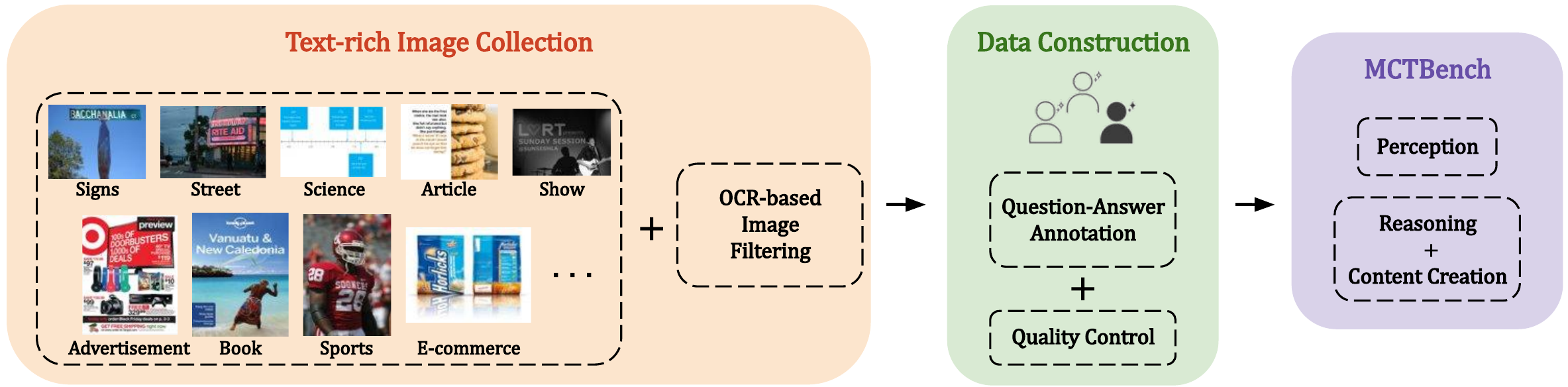}
    \caption{
    The pipeline of constructing MCTBench.
    }
    \label{MCTBench_pipline}
\end{figure}
\subsection{Overview}
\label{overview}
The MCTBench is designed to evaluate the cognitive capabilities of Multimodal Large Language Models (MLLMs) towards text-rich visual scenes. To construct the comprehensive and diverse benchmark, we collected 5,194 images from a variety of public datasets, encompassing a wide array of text-rich scenes such as natural environments, books, scientific contexts, advertisements, e-commerce, and video shots.
We meticulously annotate these images with a total of 8.5k question-answer pairs categorized into three tasks: perception, reasoning, and content creation. Specifically, MCTBench consists of 2,734 perception multiple-choice samples, 2,602 reasoning multiple-choice samples, and 3,130 content-creation samples. Figure \ref{MCTBench_pipline} illustrates the overall construction pipeline of MCTBench.
\subsection{Text-rich Images Collection} 
\label{DataCollection}
\paragraph{Image Source} The images of MCTBench are collected from 10 different publicly available datasets, aiming to incorporate comprehensive visual scenes to evaluate the cognition of Multimodal Large Language Models (MLLMs). We begin with sampling common general natural scenes (e.g., street views, competitions, road signs) from the COCO \cite{lin2015microsoft}, Flickr30k \cite{flickr30k}, GQA \cite{hudson2019gqa}, SeedBench \cite{li2023seedbench} and Visual Genome \cite{krishna2016visual} datasets.

Furthermore, we select conventional text-rich multimodal datasets: OCR-VQA \cite{mishraICDAR19}, and VizWiz \cite{gurari2018vizwiz} taken by blind photographers. 
To further diversify MCTBench, we incorporate three domain-specific scenes with broad potential applications: advertising (AutoUnderAds \cite{Hussain2017AutomaticUO}), e-commerce (FoodLogoDet-1500 \cite{Hou_2021}), and science (ScienceQA \cite{lu2022learn}). We randomly extract one frame from each video in the AutoUnderAds dataset. All data sources are specifically selected from the testsets. We adhere to the original licenses stated by all datasets.
\textbf{OCR-based image filtering}. We select the text-rich images from the sourced images, which are guided by the following guidelines.
To maintain the clarity and substantive textual content, we only retain images with valid OCR-recognized characters (with recognition probabilities higher than 0.2) of at least 10 characters.

To ensure text contributes to overall visual semantics, we select images where text regions occupy more than 10\% of the image area, after validating the impact of valid text lines on semantic expression.
These meticulous selection criteria resulted in a curated collection of high-quality and crystal-clear text-rich images, designed to challenge and inspire advancements in perceptual and cognitive understanding within textual domains.

\begin{figure}
    \centering 
    \includegraphics[width=\textwidth]{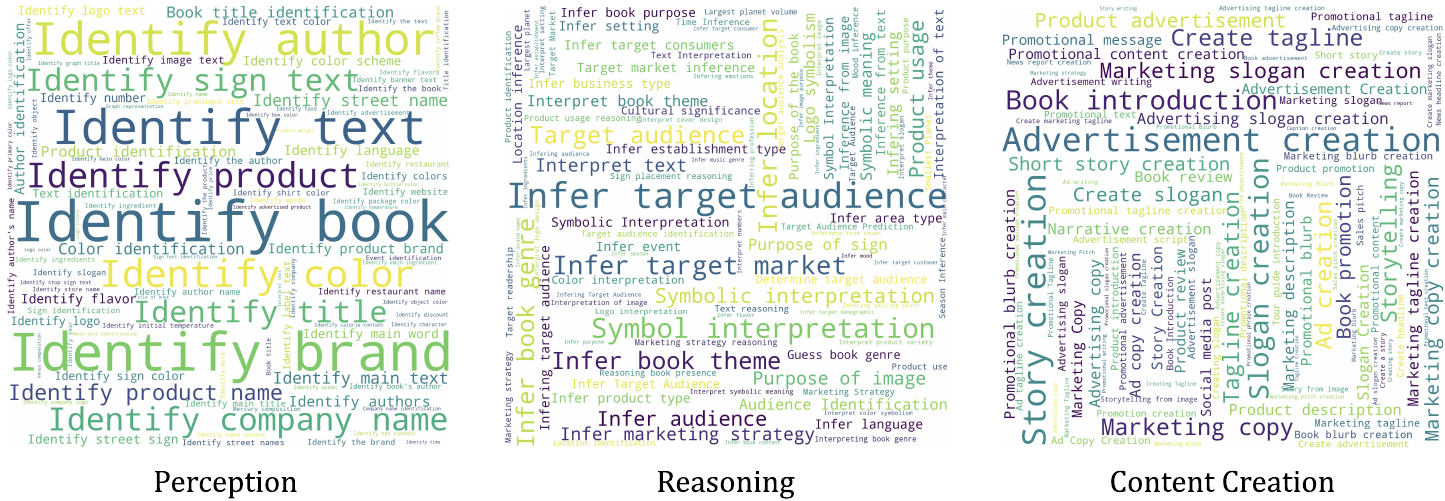}
    \caption{Visualization of the question for three different tasks using word clouds. In the word cloud, the size of a word indicates how frequently it appears. Best viewed in color.}
    \label{statics}
\end{figure}

\label{DataConstruction}
\paragraph{Annotation} Considering the bias and efficiency of the manual annotation, we employ a GPT-aided approach to generate at least 10 pseudo-questions for each image, and ask annotators to remove low-quality ones. 
All answers are human-annotated with two rounds: 
(1) Each image with at least 10 GPT-aided pseudo-questions, is randomly assigned to three annotators. Each annotator independently annotates the questions and provides answers.
(2) Quality checkers will review the annotation in the first round. If any question or image quality does not meet our annotation guidelines, the set is re-annotated by the corresponding annotators, who also revise their answers. 
Annotators in the second round are required to have at least 2 years of experience in text-rich multimodal scene annotation.
To reach an agreement, we use majority voting to determine the final answer. If majority voting does not reach an agreement (i.e., all three answers are inconsistent), we check if the discrepancy originated in the second round. If so, the question is re-annotated; if inconsistencies persist, it is discarded. The question is discarded if the discrepancy does not arise in the second round.
In addition to content-creation tasks, due to the inherent diversity of responses, we do not provide unified answers. Instead, we offer standard references generated by powerful MLLMs (e.g., GPT-4V) and meticulously reviewed by humans. 
\paragraph{Quality control} During the image quality assessment, annotators remove low-quality images (e.g., blurry, unclear text, inappropriate, solely tables/documents, and only watermark). For the QA quality assessment, annotators eliminate low-quality questions (e.g., ambiguous, overly generalized, too simplistic) and check the correctness of annotated answers (e.g., logical errors).  On the other hand, We filter out multiple-choice options with more than 30\% word count disparity and remove the questions that GPT-4 refuses to answer due to ethical concerns. To detail the specific issues addressed by each type, we assigned a label to each question and visualized the terms in Figure \ref{statics}.

\subsection{Data Construction}
\label{gen_inst}

\section {Experiments}
\label{Experiments}
In this section, we conduct comprehensive experiments to evaluate current MLLMs. Firstly, we outline experimental settings for the evaluated models and metrics used on MCTBench in Section \ref{Settings} and \ref{Evaluation}. The results obtained from these experiments allow us to perform analysis of the selected models across three categories of tasks in Section \ref{result}. Furthermore, we conduct a case study to investigate performance variations among diverse models in Section \ref{casessection}.
\subsection{Models}
\label{Settings}
To evaluate the performance of various MLLMs on MCTBench, we select a diverse range of models categorized into two primary types: \textbf{general MLLMs} and \textbf{text-enhanced MLLMs} (specifically optimized for text recognition in images). We establish two naive baselines (random choice and frequency choice) as reference points to ensure the robustness and validity of the dataset in Table \ref{tab:evaluation_all_results}. Random choice involves selecting an option at random as the prediction result, while frequency choice involves selecting the prediction result based on the option with the highest proportion in the ground truth.
\paragraph{General MLLMs} The experiment initially evaluates popular closed-source models, specifically Gemini-Pro \cite{team2023gemini} and GPT-4V(ision) \cite{Achiam2023GPT4TR}. For open-source models, we select several notable general-purpose MLLMs, including Sharegpt4V \cite{chen2023sharegpt4v}, Honeybee \cite{cha2023honeybee}, LLaVA \cite{liu2024llavanext,liu2023llava}, Otter \cite{Li2023OtterAM}, Yi-VL \cite{ai2024yi}, Qwen-VL-chat \cite{bai2023qwen} and Deepseek-VL \cite{lu2024deepseekvl} as competitive baselines. Additionally, we incorporate CogVLM \cite{wang2023cogvlm} and SPHINX-v2 \cite{lin2023sphinx}, which are enhanced for fine-grained understanding. To assess differentiated performance, we also integrate a larger open-source model, Mini-Gemini \cite{li2024minigemini}.
\paragraph{Text-enhanced MLLMs} Recent researchers propose remarkable works to tackle the understanding of text-rich images via enhancing the textual capabilities. Consequently, we select models mPLUG-DocOwl \cite{ye2023mplugowl2}, Monkey \cite{liu2024textmonkey,li2023monkey}, InternLM-XComposer2-VL \cite{internlmxcomposer2}, CogAgent \cite{hong2023cogagent} and LLaVA-NeXT \cite{liu2024llavanext} which have demonstrated strong OCR capabilities in previous evaluations. 
\subsection{Metrics}
\label{Evaluation}
MCTBench is constructed with three categories of tasks: perception, reasoning, and content-creation. Due to the standard answer provided in perception and reasoning tasks, each QA pair is designed in a multiple-choice format. In contrast, the content-creation task is considered as an open-ended generation problem due to the diversity of responses.

\paragraph{Perception and reasoning}
Perception and reasoning tasks entail the acquisition of information from input data, comprehension of images and text, and derivation of conclusions. Consequently, employing multiple-choice question-answering can effectively validate the corresponding capabilities of MLLMs, following \cite{liu2024mmbench,li2023seedbench}.
In practice, we prioritize original prompts employed by MLLMs and, if not specified, use those prompts which yield optimal results. For lengthy responses, we use regular expressions and supplementary rules to extract the option answers.
We use mean accuracy to evaluate MLLMs' perception and reasoning capability.

\paragraph{Content-creation}
To ensure consistency and efficiency in evaluation, we implement automatic evaluation for content-creation tasks. However, due to the diversity of answers in content creation, standard responses are not feasible, leading us to establish competitive references. 
These references are generated through manually crafted responses from text-only GPT-4, based on inputs from OCR recognition and detailed descriptions.
The evaluation is grounded on four principal aspects: relevance, faithfulness, creativity, and instruction following. Subsequently, we employ machines (e.g., GPT-4V) to compare other MLLMs against our references with the mentioned principal, and categorize their performance as Good, Same, or Bad (i.e., the GSB metric). We measure each model's performance by calculating the percentage of 'Good' and 'Same' ratings relative to all questions, indicating how many outperform or match the constructed reference (i.e., $(G+S)/(G+S+B)$).
To assess correlations and validate the reliability of machine evaluation, we also conduct a manual evaluation on subsets of the data using the mentioned metrics (GSB) and compare them with machine evaluation. We integrate three powerful MLLMs (GPT-4V \cite{Achiam2023GPT4TR}, Gemini-Pro \cite{team2023gemini} and LLaVA-NeXT \cite{liu2024llavanext}) as evaluators. We measure the evaluation correlation between humans and machines using accuracy and Pearson correlation \cite{benesty2009pearson} on GSB scores. Table \ref{tab:correlation} illustrates the results between three machine evaluators and human evaluations, indicating that GPT-4V achieves a top correlation score.

\subsection{Results}
\label{result}

\begin{table}[t!]%
\centering
\setlength\tabcolsep{2.5pt}
\begin{tabular}{ll|ccc|ccc}
\toprule
 & \multicolumn{1}{c}{} & \multicolumn{1}{c}{} & \multicolumn{2}{c}{Cognition} & \multicolumn{3}{c}{Average Scores} \\ \cline{4-8} 
\multirow{-2}{*}{Model} & \multicolumn{1}{c}{\multirow{-2}{*}{Params}} & \multicolumn{1}{c}{\multirow{-2}{*}{\begin{tabular}[c]{@{}c@{}}Perception\\ \end{tabular}}} & \multicolumn{1}{c}{\begin{tabular}[c]{@{}c@{}}Reasoning\\ \end{tabular}} & \multicolumn{1}{c}{\begin{tabular}[c]{@{}l@{}}Content-\\Creation${}^{*}$ \end{tabular}} & \multicolumn{1}{c}{MC} & \multicolumn{1}{c}{Cog} & \multicolumn{1}{c}{All} \\ \hline
\multicolumn{8}{c}{Naive Baseline} \\ \hline
{\color[HTML]{939393} Random choice} & {\color[HTML]{939393} -} & {\color[HTML]{939393} 25.00} & {\color[HTML]{939393} 25.00} & {\color[HTML]{939393} -} & {\color[HTML]{939393} -} & {\color[HTML]{939393} -} & {\color[HTML]{939393} -} \\
{\color[HTML]{939393} Frequency choice} & {\color[HTML]{939393} -} & {\color[HTML]{939393} 25.16} & {\color[HTML]{939393} 25.52} & {\color[HTML]{939393} -} & {\color[HTML]{939393} -} & {\color[HTML]{939393} -} & {\color[HTML]{939393} -} \\ \hline
\multicolumn{8}{c}{General MLLMs} \\ \hline
GPT-4V \cite{Achiam2023GPT4TR} & - & \underline{83.58} & \textbf{74.21} & \textbf{87.35} & \textbf{78.90} & \textbf{83.12} & \textbf{81.71} \\
Gemini-Pro \cite{team2023gemini} & - & 78.79 & 70.18 & 56.78 & 74.49 & 65.63 & 68.58 \\
Yi-VL \cite{ai2024yi} & 6B & 77.25 & 72.33 & 41.45 & 74.79 & 58.12 & 63.68 \\
Deepseek-VL \cite{lu2024deepseekvl} & 7B & 76.74 & 68.79 & 57.25 & 72.77 & 65.01 & 67.59 \\
Honeybee \cite{cha2023honeybee} & 7B & 72.60 & 67.22 & 73.64 & 69.91 & 71.78 & 71.15 \\
Otter \cite{Li2023OtterAM} & 7B & 58.12 & 54.42 & 31.70 & 56.27 & 43.99 & 48.08 \\
Qwen-VL-chat \cite{bai2023qwen} & 7B & 77.98 & 70.68 & 67.53 & 74.33 & 70.93 & 72.06 \\
Sharegpt4V \cite{chen2023sharegpt4v} & 13B & 74.54 & 69.49 & 66.19 & 72.02 & 69.10 & 70.07 \\
LLaVA-1.5 \cite{liu2023llava} & 13B & 78.09 & 72.56 & 66.47 & 75.33 & 70.90 & 72.37 \\
SPHINX-v2 \cite{lin2023sphinx} & 13B & 78.02 & 71.94 & 62.30 & 74.98 & 68.64 & 70.75 \\
CogVLM \cite{wang2023cogvlm} & 17B & 71.40 & 69.52 & 65.61 & 70.46 & 68.04 & 68.84 \\
Mini-Gemini \cite{li2024minigemini} & 34B & \textbf{83.83} & \underline{73.33} & \underline{86.76} & \underline{78.58} & \underline{82.67} & \underline{81.31} \\ \hline
\multicolumn{8}{c}{Text-enhanced MLLMs} \\ \hline
IXC 2 \cite{internlmxcomposer2} & 7B & 78.05 & \underline{72.10} & \underline{74.45} & 75.08 & \underline{74.76} & \underline{74.87} \\
Monkey \cite{li2023monkey} & 7B & \underline{79.22} & \textbf{72.64} & 59.56 & \underline{75.93} & 67.75 & 70.47 \\
TextMonkey \cite{liu2024textmonkey} & 7B & 71.80 & 69.45 & 22.81 & 70.63 & 46.72 & 54.69 \\
mPLUG-DocOwl \cite{ye2023mplugdocowl} & 10B & 75.05 & 70.06 & 60.87 & 72.56 & 66.71 & 68.66 \\
CogAgent \cite{hong2023cogagent} & 34B & 58.56 & 56.46 & 56.86 & 57.51 & 57.19 & 57.29 \\
LLaVA-NeXT \cite{liu2024llavanext} & 34B & \textbf{83.87} & 71.64 & \textbf{85.30} & \textbf{77.76} & \textbf{81.53} & \textbf{80.27} \\ \hline
\end{tabular}
\caption{Evaluation results for MLLMs on MCTBench. \textbf{MC} means the average score of the two tasks (perception and reasoning) in multiple-choice format. \textbf{Cog} means the average scores of the two cognitive tasks (reasoning and content-creation). \textbf{All} means the overall average scores of all tasks. ${}^{*}$The content-creation task is scored using the percentage of 'Good' and 'Same' ratings by the GSB metric except for accuracy used in Perception and reasoning tasks. Numbers in \textbf{Bold} and \underline{underline} represent the top-2 results in each task.}
\label{tab:evaluation_all_results}
\end{table}

\paragraph{Perception}
Firstly, experiments are conducted to verify the perceptual performances of each model as baselines. As shown in Table~\ref{tab:evaluation_all_results}, most models achieved satisfying scores on the perception task. Closed-source models (e.g., GPT-4V \cite{Achiam2023GPT4TR}) demonstrated excellent accuracy, while some open-source models (e.g., Mini-Gemini \cite{li2024minigemini}) demonstrated superior perception capabilities, surpassing the their performance. Models with higher resolutions and more parameters (e.g., LLaVA-NeXT \cite{liu2024llavanext} and Mini-Gemini \cite{li2024minigemini}), typically performed better. Among similarly-sized models, text-enhanced MLLMs generally outperformed others by effectively extracting text from images and generating precise responses.
\paragraph{Reasoning}
The reasoning task is more challenging than perception. 
It requires not only effective extraction and fusion of visual and textual features, but also involves comprehensive inference to generate accurate responses. 
As shown in Table~\ref{tab:evaluation_all_results}, most models exhibited a significant drop in scores due to the increased difficulty of reasoning tasks compared to perception tasks.

Notably, GPT-4V demonstrates exceptional performance among MLLMs, surpassing most models by a significant margin.
Besides, there still exists a positive correlation between a model's performance and the number of its parameters. 
This phenomenon arises from larger models' enhanced ability, to comprehend text and integrate image information for reasoning more effectively. 

Nevertheless, text-enhanced MLLMs have not substantially outperformed general models on reasoning tasks. 
Given that text-enhanced models have achieved superior results in perception tasks, we posit their effectiveness in recognizing text within images. 
However, achieving higher scores in reasoning tasks necessitates the ability to analyse and summarise effectively.
TextMonkey \cite{liu2024textmonkey} shows the least performance gap and achieves results comparable to the perception task. However, most text-enhanced models are not explicitly trained in this aspect, and consequently do not outperform general MLLMs.

\begin{table}[ht]
    \centering
\begin{tabular}{lccc}
\toprule
 & \multirow{1}{*}{GPT-4V} & \multirow{1}{*}{Gemini-Pro} & {\makecell{Best open-source\\(LLaVA-NeXT)}} \\ \hline
Accuracy       &     79.38   &  70.71  &   65.22     \\
Pearson Correlation  &  0.558   &  0.380  &  0.304    \\ \hline
\end{tabular}
\caption{Correlation analysis between automatic machine and human evaluation on content-creation using the accuracy and Pearson correlation coefficient.}
\label{tab:correlation}
\end{table}

\paragraph{Content-creation}
The open-ended creation task differs from the aforementioned tasks, leading to the observation of a broader array of perspectives. 
As mentioned in Section \ref{Evaluation}, the evaluation is based on four principal aspects: relevance, faithfulness, creativity, and instruction following.
Therefore, as the content-creation task emphasizes the generation of suitable text for images, we also notice significantly enhancement on performances by employing larger language models. 
However, the performance of text-enhanced MLLMs exhibits considerable variation due to differences in training objectives. 
Some models are specifically trained to extract structured text information or comprehend lengthy text inputs. 
Consequently, they may underperform relative to general models when tasked with creative endeavours.
We also found that the closed-source model Gemini-Pro \cite{team2023gemini} is unable to achieve good results. 
A notable discrepancy in its scores of template following demonstrates that Gemini struggles to create corresponding formats for task specifications.
For instance, in the task of generating a slogan, Gemini may tend to produce lengthy paragraphs instead of concise phrases. 
Some text-enhanced MLLMs, such as TextMonkey, also have similar phenomena.

\paragraph{Summary}
MLLMs have shown certain multi-modal capabilities in perception tasks, with mainstream models achieving commendable performances. However, in reasoning tasks, existing models still have room for improvement. Performance on content-creation tasks indicates that text-enhanced MLLMs trained for different types of tasks may lose some creative capabilities. We suggest that reasoning and creation tasks serve as distinct dimensions for evaluation, offering insights into the model's comprehension of input and its proficiency in generating output responses. While existing models excel in basic perception tasks, achieving comparable competence in reasoning and creation tasks remains challenging.

\begin{table}[!t]
    \centering
    \begin{tabular}{l| c c c}
    \hline
        Input & Perception & Reasoning & Creation \\ \hline
        Image + OCR texts & 75.60 & 71.68 & 65.45 \\ 
        Image (text regions removed) & 62.22 & 62.22 & 46.89 \\ 
        Image (text regions only) & 74.69 & 68.01 & 58.78 \\ 
        Image (baseline) & 78.09 & 72.56 & 66.47 \\ \hline
    \end{tabular}
    \caption{The effect of visual and textual information.}
    \label{ablation}
\end{table}

\subsection{Ablation Study}
\paragraph{The effect of visual and textual information. }
In Table~\ref{ablation}, we conduct several experiments to reveal the effect of visual and textual information in images of our benchmark. 
We use the LLaVA-1.5 as the baseline.
Firstly, we conduct an experiment that adds OCR texts as input. 
As shown in the table, even if the OCR texts are contained as input, the model cannot gain explicit improvement. This demonstrates that the MCTBench does not rely solely on texts to get answers. 
Furthermore, we conduct an experiment that removes all texts from images to check the importance of texts in our benchmark. 
Specifically, we detected and blurred all the texts in images. We noticed that the performance dropped significantly. This also demonstrates that it is difficult for a model to correctly answer the question without text in the image.
Additionally, if we only keep the text region, and delete other background parts in the image. The performance is between the above two experiments, indicating that MCTBench relies both on textual and visual information, to get the final result. To conclude, explicitly adding OCR texts, or removing text/image parts does not help, or even lead to worse performance on MCTBench. MLLMs are required to jointly recognize related textual and visual patterns, to answer the questions correctly.

\begin{figure}
    \centering
\includegraphics[width=\textwidth]{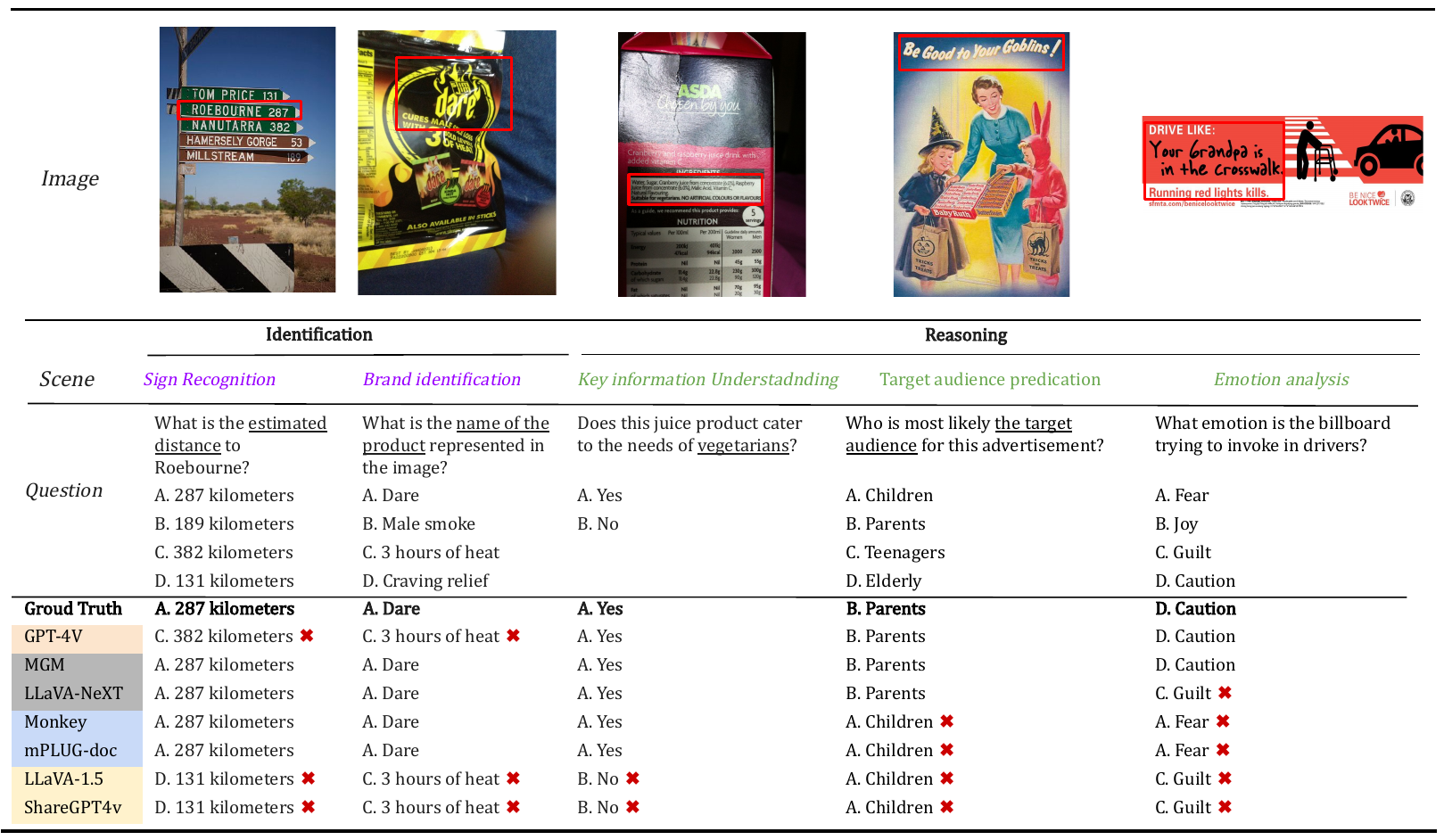}
    \caption{The cases of predication from different MLLMs divided into four groups: GPT-4V \cite{Achiam2023GPT4TR}, Mini-Gemini \cite{li2024minigemini}(MGM) and LLaVA-NeXT \cite{liu2024llavanext} for larger model size, Monkey \cite{li2023monkey} and mPLUG-DocOwl \cite{ye2023mplugdocowl} for text-enhanced MLLMs, LLaVA-1.5 \cite{liu2023improvedllava} and ShareGPT4V \cite{chen2023sharegpt4v} for the general MLLMs}
    \label{CaseStudy1}
    \label{fig:case_study_1}
\end{figure}

\begin{figure}
    \centering
\includegraphics[width=\textwidth]{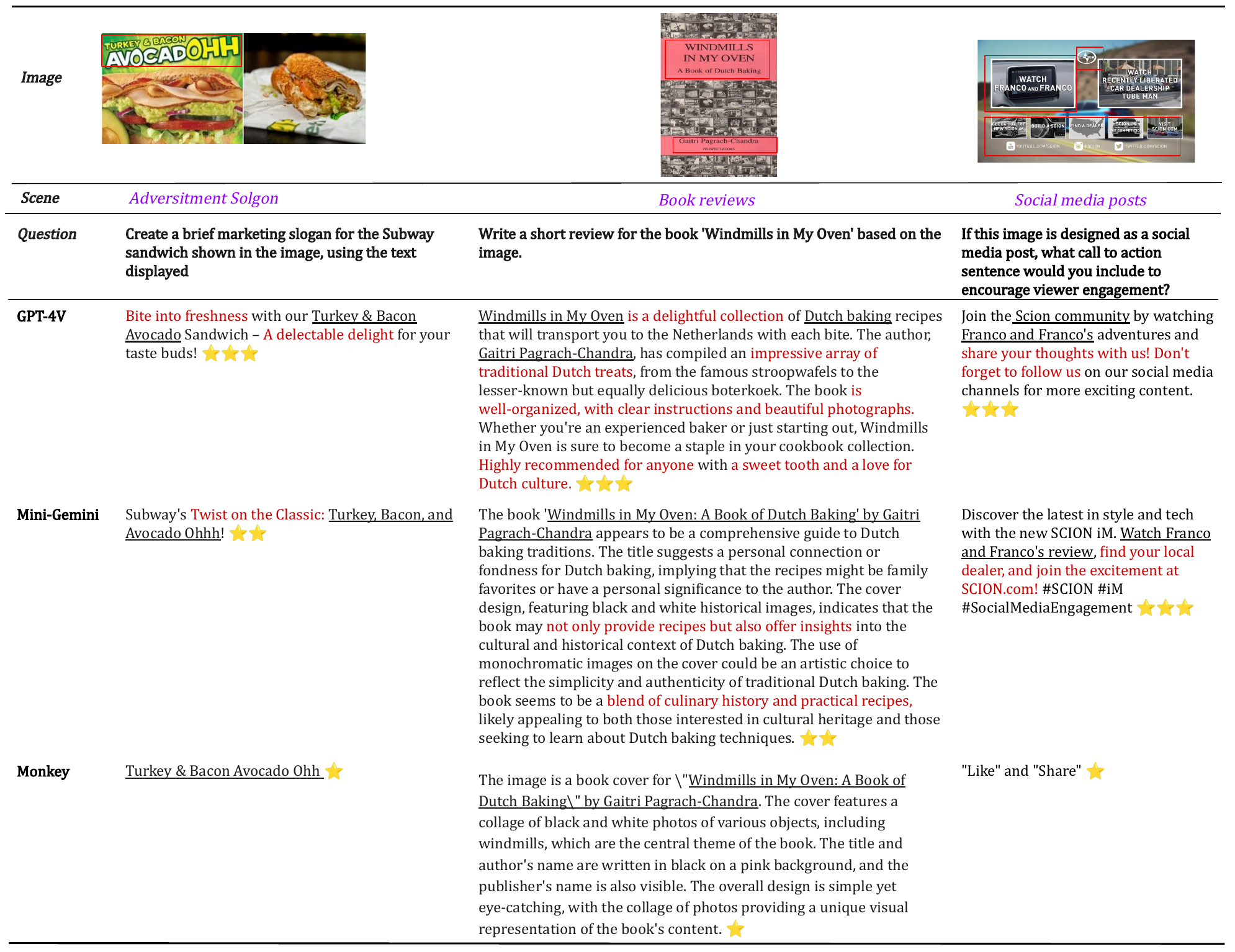}
    \caption{The cases of predication on content-creation tasks from three representative MLLMs: GPT-4V \cite{Achiam2023GPT4TR}, Mini-Gemini \cite{li2024minigemini} and  Monkey \cite{li2023monkey}. We mark high-quality sentences in red, words hit the text in the image with underlining, and rate the quality of the generation with stars.}
    \label{CaseStudy2}
    \label{fig:case_study_2}
\end{figure}

\subsection{Case Study}
\label{casessection}
For the perception and reasoning tasks, we select representative cases shown in Figure \ref{CaseStudy1}. Specifically, we select GPT-4V as a strong reference along with several open-source representative MLLMs, split into three groups: large model size and resolution MLLMs (Mini-Gemini \cite{li2024minigemini} and LLaVA-NeXT \cite{liu2024llavanext}), text-enhanced MLLMs (Monkey \cite{li2023monkey} and mPLUG-DocOwl \cite{ye2023mplugdocowl}), and general MLLMs (LLaVA-1.5 \cite{liu2023improvedllava} and ShareGPT4V \cite{chen2023sharegpt4v}). 
For the selected perception question, both high-resolution MLLMs and text-enhanced MLLMs perform well, while general MLLMs fail in fine-grained understanding. On the contrary, text-enhanced MLLMs excelled in perception but performed poorly in reasoning tasks. Larger models like Mini-Gemini \cite{li2024minigemini}, LLaVA-NeXT \cite{liu2024llavanext}, and GPT-4V \cite{Achiam2023GPT4TR} handle reasoning better by effectively integrating textual and visual elements.
 
For the content-creation task, using GPT-4V \cite{Achiam2023GPT4TR} as a robust reference, we select general and text-enhanced models that performed well on MCTBench, and conduct case studies in three scenarios. As shown in Figure \ref{CaseStudy2}, GPT-4V \cite{Achiam2023GPT4TR} significantly surpassed other models in content creation quality. Mini-Gemini \cite{li2024minigemini} also showed consistent performance across cognitive tasks, while text-enhanced models like Monkey \cite{li2023monkey} were limited to text recognition and basic descriptions, resulting in less attractive content.

\section{Limitations}
\label{limitaion}
Our dataset primarily focuses on English, which may limit the generalization of our findings to multilingual scenes. 
Although we believe that the cognitive capacities of MLLMs should theoretically extend to other languages, we have not empirically substantiated this assertion in the present study.
Additionally, we have only selected a subset of representative models for evaluation due to space constraints. This selection may not cover the full spectrum of currently available MLLMs. Our future work aims to provide evaluation results for a more extensive range of models.

\section{Conclusion}
In this work, we introduce MCTBench, a comprehensive benchmark designed to evaluate the cognitive capabilities of MLLMs in text-rich visual scenes. The MCTBench comprises 5.2k images and 8.5k question-answer pairs, covering a range of tasks including reasoning, content creation for cognitive assessment, and conventional perception. Evaluations of MLLMs on MCTBench reveal that current MLLMs still need further advancements in cognitive capabilities, despite their superior perception performance. We hope that MCTBench will motivate researchers to further improve the cognitive capabilities of MLLMs in text-rich visual scenes, thereby enhancing the practical utility of AI in real-world applications.
\bibliographystyle{unsrt}  
\bibliography{references}

\end{document}